\documentclass[times, review, 10pt]{elsarticle}
\usepackage{amssymb, amsthm, amsmath}
\usepackage{graphics, subfigure,float,wrapfig}
\usepackage{makecell}
\usepackage{algorithm, algpseudocode}
\newcommand{\Input}{\item[\textbf{Input:}]}
\newcommand{\Output}{\item[\textbf{Output:}]}
\usepackage[colorlinks,linkcolor=blue]{hyperref}

\journal{Pattern Recognition}

\begin{document}

\begin{frontmatter}

%% Title, authors and addresses

%% use the tnoteref command within \title for footnotes;
%% use the tnotetext command for theassociated footnote;
%% use the fnref command within \author or \address for footnotes;
%% use the fntext command for theassociated footnote;
%% use the corref command within \author for corresponding author footnotes;
%% use the cortext command for theassociated footnote;
%% use the ead command for the email address,
%% and the form \ead[url] for the home page:
%% \title{Title\tnoteref{label1}}
%% \tnotetext[label1]{}
%% \author{Name\corref{cor1}\fnref{label2}}
%% \ead{email address}
%% \ead[url]{home page}
%% \fntext[label2]{}
%% \cortext[cor1]{}
%% \affiliation{organization={},
%%             addressline={},
%%             city={},
%%             postcode={},
%%             state={},
%%             country={}}
%% \fntext[label3]{}

\title{TCMM: Token Constraint and Multi-Scale Memory Bank of Contrastive Learning for Unsupervised Person Re-identification}

%% use optional labels to link authors explicitly to addresses:
%% \author[label1,label2]{}
%% \affiliation[label1]{organization={},
%%             addressline={},
%%             city={},
%%             postcode={},
%%             state={},
%%             country={}}
%%
%% \affiliation[label2]{organization={},
%%             addressline={},
%%             city={},
%%             postcode={},
%%             state={},
%%             country={}}

\author[a]{Zheng-An Zhu}
\ead{cca105p@cs.ccu.edu.tw}
\author[a]{Hsin-Che Chien}
\ead{hsin@csie.io}
\author[a]{Chen-Kuo Chiang\corref{cor1}}
\ead{ckchiang@cs.ccu.edu.tw}

\affiliation[a]{
organization={Computer Science and Information Engineering, National Chung Cheng University},   
            %Department and Organization
            addressline={No.168, Sec. 1, University Rd., Minhsiung}, 
            city={Chiayi},
            postcode={621301}, 
            country={Taiwan}
            }

\cortext[cor1]{Corresponding author}

\begin{abstract}
%% Text of abstract
This paper proposes the ViT Token Constraint and Multi-scale Memory bank (TCMM) method to address the patch noises and feature inconsistency in unsupervised person re-identification works. Many excellent methods use ViT features to obtain pseudo labels and clustering prototypes, then train the model with contrastive learning. However, ViT processes images by performing patch embedding, which inevitably introduces noise in patches and may compromise the performance of the re-identification model. On the other hand, previous memory bank based contrastive methods may lead data inconsistency due to the limitation of batch size. Furthermore, existing pseudo label methods often discard outlier samples that are difficult to cluster. It sacrifices the potential value of outlier samples, leading to limited model diversity and robustness. This paper introduces the ViT Token Constraint to mitigate the damage caused by patch noises to the ViT architecture. The proposed Multi-scale Memory enhances the exploration of outlier samples and maintains feature consistency. Experimental results demonstrate that our system achieves state-of-the-art performance on common benchmarks. The project is available at \href{https://github.com/andy412510/TCMM}{https://github.com/andy412510/TCMM}.

\end{abstract}

%Graphical abstract
% \begin{graphicalabstract}
% \includegraphics[width=\linewidth]{figures/Graphical_abstract.png}
% \end{graphicalabstract}

%%Research highlights
% \begin{highlights}
% \item The ViT token constraint reduces patch noises on the self-attention mechanism.
% \item The prototype memory module alleviates noise and feature inconsistency issues.
% \item The instance memory module fully leverages the value of all training data.
% \item Experimental results demonstrate that proposed TCMM achieves SOTA performance.
% \end{highlights}

\begin{keyword}
%% keywords here, in the form: keyword \sep keyword
Unsupervised Person Re-identification \sep Vision Transformer \sep Pseudo labels \sep Prototypes \sep Contrastive Learning.
%% PACS codes here, in the form: \PACS code \sep code

%% MSC codes here, in the form: \MSC code \sep code
%% or \MSC[2008] code \sep code (2000 is the default)

\end{keyword}

\end{frontmatter}

%% \linenumbers
%% The Appendices part is started with the command \appendix;
%% appendix sections are then done as normal sections
%% \appendix

\section{Introduction} 
\label{intro}
This paper proposes multiple modules based on contrastive learning to mitigate patch noises and feature inconsistency that may arise in unsupervised person re-identifi-
cation (re-id) tasks. 
Unsupervised person re-id has wide applications in the real world, such as city security monitoring, traffic control and management, event investigation and crime tracking. In real-world surveillance systems, a large amount of video data is generated daily. The application of supervised learning on this data is limited by the amount of labeled data. Unsupervised learning can more effectively handle and utilize this vast amount of data, improving the scalability and practicality of the models.
Previous works on person re-id have utilized ViT models to extract discriminative re-id features~\cite{NFormer, fed}, or employed pseudo labels and prototypes to aid in unsupervised training~\cite{MaskPre, AFC, ise, pplr, secret, HSP-MFL}. Meanwhile, many excellent unsupervised person re-id systems have incorporated these technologies and trained the model using contrastive learning~\cite{rethinking, ACFL, CCIOL, pass, spcl, clusternce, mcrn, dcmip}.

ViT effectively captures global context information through self-attention mechanism, which enables it to attend to all parts of an image simultaneously. This global context understanding is crucial for unsupervised learning tasks, where the model needs to extract meaningful features from images without relying on class labels or localized information.
ViT is built upon treating image patches as "visual tokens", utilizing patch embedding and learning patch-to-patch attention mechanisms throughout the process. Unlike textual tokens provided in natural language processing, visual tokens must be learned initially and refined iteratively to learn ViTs more effectively.

Despite the remarkable success of ViT in computer vision~\cite{vit}, the presence of patch noises in ViT can significantly impact its performance and effectiveness. Patch noises refer to the situation where the patches extracted from an image contain irrelevant or distracting information, such as background clutter, texture variations, or unrelated objects.
The self-attention mechanism has a particular ability to ignore or assign lower attention weights to patch noises, but this is not always sufficient. When calculating attention weights, all patches participate in the computation, and even if they are assigned lower weights, the cumulative effect can still impact the overall results. Additionally, patch noises increase the computational burden, dilute useful information, and cause contextual confusion, all of which affect the effectiveness of self-attention. Therefore, although the self-attention mechanism can theoretically handle patch noises, these factors still affect the model performance in practical applications. As a result, the model may struggle to differentiate between relevant and irrelevant features, leading to decreased accuracy and robustness in downstream tasks.
RSPC~\cite{rspc} demonstrates that the vulnerability of modern state-of-the-art Transformers stems from the inherent sensitivity of the self-attention mechanism to input patches. Based on patch-based input, the self-attention mechanism is often overly sensitive to noise in each patch.
In summary, patch noises introduce noise information into the input sequence, undermining its ability to learn meaningful representations and perform effectively in unsupervised person re-id tasks. Addressing patch noise is crucial for improving the performance and robustness of ViT.

Pseudo labels effectively bridge the gap between unsupervised and supervised learning paradigms by providing a form of weak supervision. While pseudo labels may not be as accurate as ground truth labels, pseudo labels still convey valuable information about the underlying structure of the data. By incorporating pseudo labels during training, unsupervised person re-id models can benefit from labeled and unlabeled data, leading to better generalization and performance. 
Although pseudo label techniques have significantly improved performance in unsupervised learning, they share a common drawback: errors often affect predictions on unlabeled images. Using these predictions for supervised learning can lead to confirmation bias towards errors, resulting in a deteriorated model. Some existing pseudo label-based methods address this issue by manually adjusting thresholds to select uncertain predictions to be discarded. However, their performance largely depends on manually adjusted thresholds~\cite{dcloss, hl_consis}.
Some methods employ additional error correction teacher networks~\cite{pplr} to refine the generated pseudo labels. However, using multiple teacher networks incurs significant extra costs.
On the other hand, some methods choose to completely discard unclassified outlier samples and adopt a memory bank to store features or cluster prototypes~\cite{pass, clusternce, eln}. ELN~\cite{eln} further demonstrates that ignoring invalid pseudo labels instead of correction can alleviate confirmation bias and learn accurate models. 
However, outlier samples may represent valuable yet difficult-to-classify samples. Typically, many outliers are generated early in training. Discarding these outliers may harm the final performance~\cite{spcl}, limiting the diversity and robustness of the model. 
Moreover, previous methods often use a memory bank to store instance features and then use all instance features to train contrastive learning. Due to the limitation of batch size, only a few samples are updated in each iteration. This can lead data inconsistency between these samples and the ones that should be very similar.

This research presents a ViT Token Constraint and Multi-scale Memory bank architecture (TCMM) that focuses on solving the downstream unsupervised person re-id task.
In this paper, the ViT token constraint is proposed to tackle the vision transformer patch noises. The multi-scale memory bank is designed to alleviate the instance consistency and outlier samples issues in existing state-of-the-art unsupervised person re-id works.
ViT token constraint adopts the concept of contrastive learning to bring the discriminative patch feature closer to the image feature while pushing patch noises away from the image feature, thereby limiting the negative impact of patch noises on the model.

\begin{figure}[t]
\centering
\subfigtopskip=10pt  % 第一個子圖跟 caption 文字的距離
\label{fig:hierarchical}

\subfigure[Instance level architecture]{
\label{fig:hierarchical_ins}
\includegraphics[width=.3\linewidth]{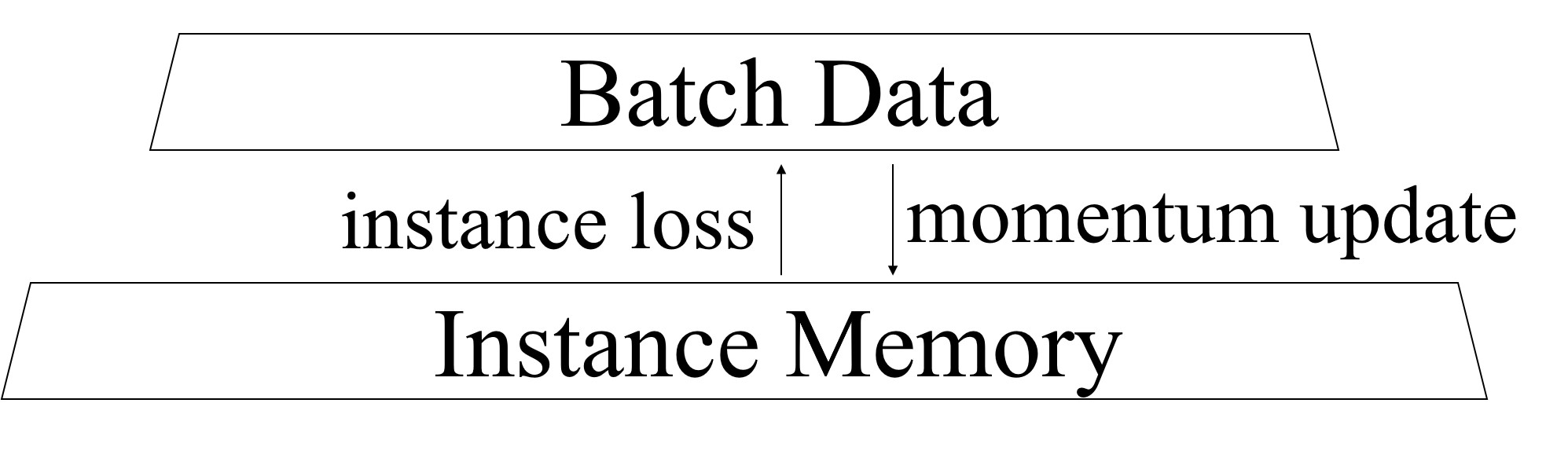}
}
\subfigure[Prototype level architecture]{
\label{fig:hierarchical_pro}
\includegraphics[width=.3\linewidth]{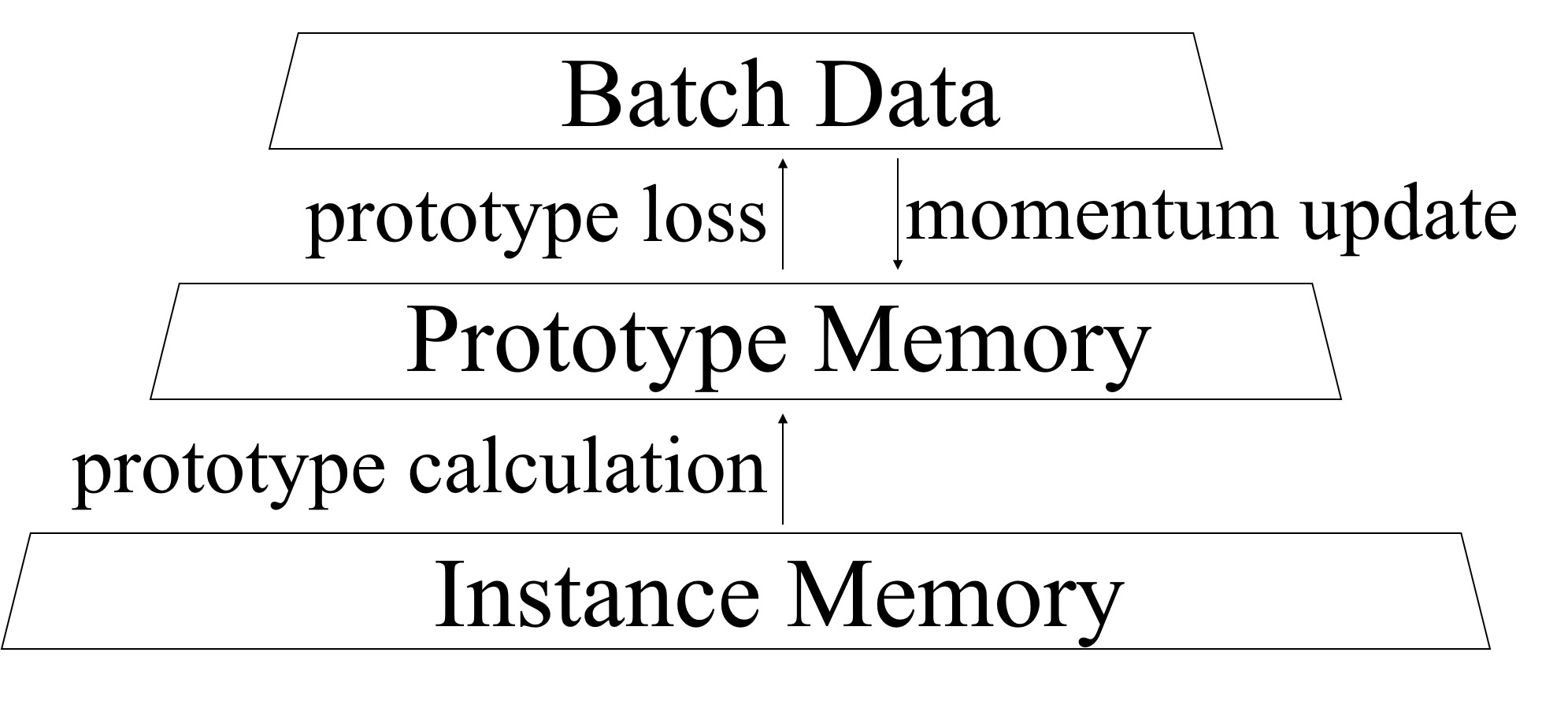}
}
\subfigure[TCMM architecture]{
\label{fig:hierarchical_our}
\includegraphics[width=.3\linewidth]{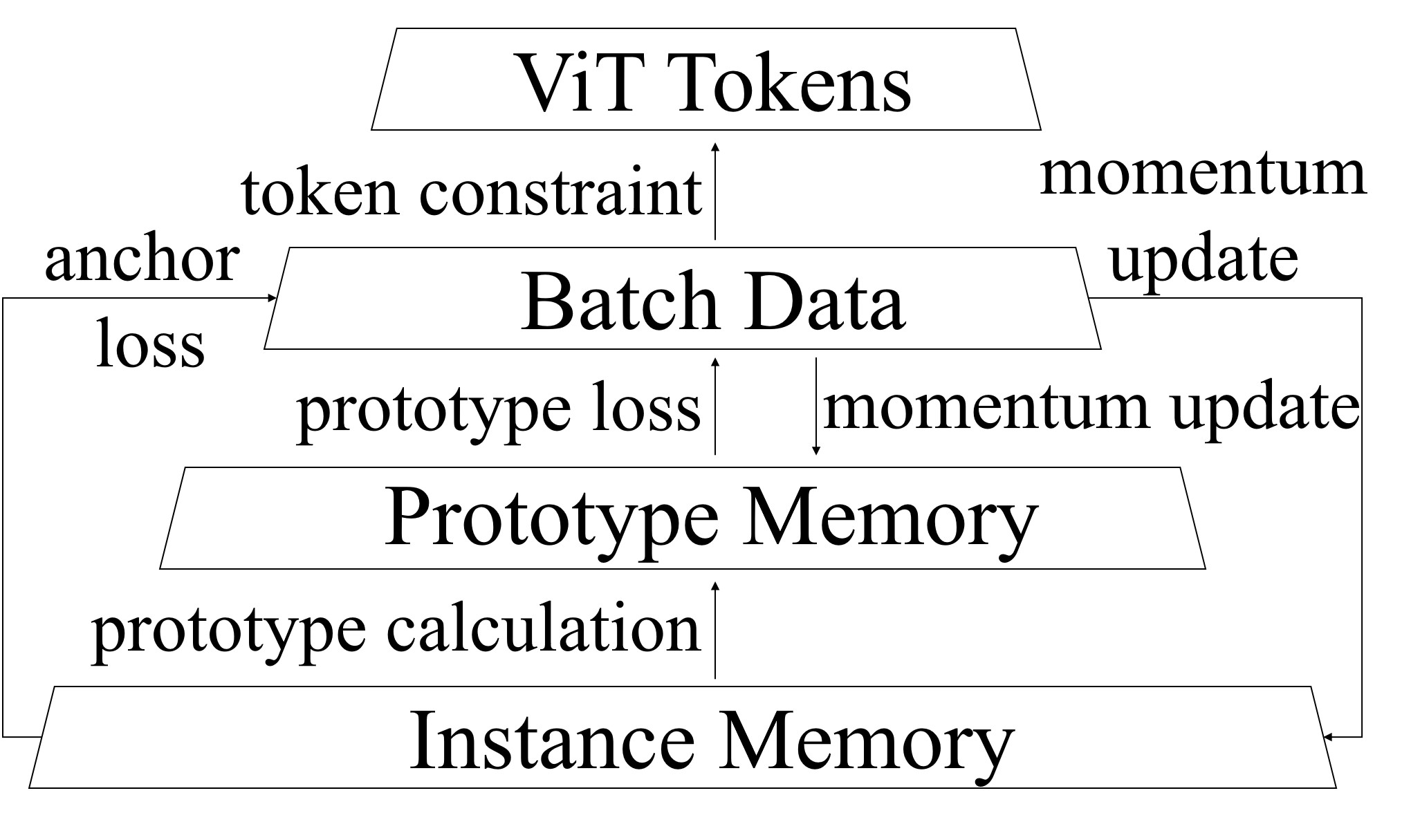}
}
\caption[Hierarchical conceptual comparison.]{Hierarchical conceptual comparison of different feature level memory. \ref{fig:hierarchical_ins} shows instance memory composed of all data, including outliers, and contrastive learning training with batch data. \ref{fig:hierarchical_pro} shows the calculation of cluster prototype memory using data without outliers and contrastive learning training with batch data. \ref{fig:hierarchical_our} is the TCMM architecture. TCMM first applies token constraint to ViT features to reduce the interference of patch noise on the model. TCMM uses prototype memory without outliers to mitigate the negative impact of potential noise on the model and address feature inconsistency. TCMM also considers instance memory that includes outliers to fully account for the value of all samples. Through these mechanisms, TCMM achieves state-of-the-art performance without adding an additional correcting model or changing the pseudo-label generation process and results.}
\end{figure}

The multi-scale memory bank aims to mitigate the issues of feature inconsistency and explore the potential value of outlier samples without increasing additional model costs or altering the pseudo label generation process and result. 
This study introduces a prototype level memory module, leveraging prototype memory and prototype contrastive loss to address the issue of feature inconsistency and enhance the model ability to distinguish between different clusters. Calculating and using cluster prototype features can make the feature representation in the memory bank more stable and consistent than sample features. The instance features update in one iteration may not be consistent with the features of the same cluster due to the limitations imposed of data imbalance. This means that the feature representations within the same cluster are generated by models in different states at different iterations, leading to feature inconsistency. The stability and consistency of the feature representation can be ensured by treating the centroid as the cluster feature and focusing on the representative feature of each cluster. The prototypes level memory module utilizes prototypes from the same cluster as positive samples and prototypes from other clusters as negative samples, thereby avoiding feature inconsistency and making the model more robust and discriminative. This paper also proposes a novel instance level memory module that utilizes instance memory and anchor contrastive loss to explore the potential value of outlier samples. The instance level memory module considers all samples, brings closer the harder positive samples and separates the harder negative samples, encouraging the model to learn from difficult samples. Thus, it enables the model to cover various data variations and improve generalization.
The hierarchical conceptual comparison between TCMM and previous different feature level memory systems is shown in Figure \ref{fig:hierarchical}.

The contributions of this paper can be summarized as follows:
1). The TCMM method proposed in this paper combines contrastive learning and introduces the ViT token constraint and multi-scale memory bank to address the challenges in existing unsupervised person re-id tasks.
2). The proposed ViT token constraint can reduce the adverse effects of patch noises on the self-attention mechanism by constraining token features.
3). The multi-scale memory bank encourages the model to focus more on difficult samples while ensuring feature consistency, allowing the model to better differentiate between the similarities and differences among different clusters. By enabling the model to learn more discriminative re-id features, the system optimizes the model performance.
4). The experiments demonstrate that the TCMM framework achieves state-of-the-art performance without increasing additional model costs or altering the pseudo label generation process and result.

% ===========================  Related Work  ===========================%
\section{Related Work}
\label{sec:related}
% Person re-identification has received widespread attention in the computer vision field~\cite{gepp, dcmip, RTMem, HSP-MFL, CCIOL, secret, mcrn, rethinking, MaskPre}.
Person re-identification has received widespread attention in the computer vision field~\cite{gepp, SSMs, IQE, DLCL, DHCL, BSE, AAAM, DFDSN-Net, MTNet, Shah, HPcL, NFormer}. Benefiting from the powerful performance of contrastive learning, many unsupervised person re-id methods use contrastive loss to train models. In these studies based on contrastive learning, ViT features \cite{NFormer, mcrn, dcmip}, pseudo labels \cite{MaskPre, AFC, pplr, HSP-MFL}, and prototypes \cite{ise, secret, RTMem, rethinking} are often incorporated to enhance the effectiveness of contrastive loss. This section briefly introduces and summarizes the application of these techniques in previous unsupervised person re-id methods.

\subsection{Vision Transformer}
\label{sec:related-vit}

In recent years, many studies have introduced ViT into computer vision tasks with outstanding performance. When combined with the powerful feature extraction capabilities of ViT, the model can effectively learn from large amounts of unlabeled data, capturing nuanced details and diverse characteristics of the input data.
RSPC~\cite{rspc} proposed a novel training approach that improves the robustness of Transformers from a new perspective. Specifically, RSPC first identifies and masks/distorts the most vulnerable patches, and then explicitly reduce their sensitivity by aligning the intermediate features between clean and corrupted samples. The construction of patch corruption enables adversarial learning for the subsequent feature alignment process, which is particularly effective and inherently different from existing methods.
PaCa~\cite{paca} guides clustering-for-attention and attention-for-clustering in ViT models through joint learning. Given an input $X$, the lightweight cluster module first computes the cluster assignment $C$ for predefined small clusters $M$ to identify meaningful clusters. Then, through matrix multiplication between the cluster assignment $C$ and $X$, $M$ potential "visual tokens" $Z$ are formed. The clustering $C$ can directly visualizes as heatmaps to demonstrate the content captured by the trained model.
NFormer~\cite{NFormer} utilizes a Transformer model to extract relations between input person images. Specifically, NFormer employs low-rank factorization with a small number of landmarks to effectively shape the relational graph between images, reducing interference from irrelevant representations and further alleviating boundary computations.

DCMIP~\cite{dcmip} believes that the uni-proxy mechanisms of existing methods are affected by inter-class similarity and intra-class variance. DCMIP proposes discrepant cluster proxies based contrastive learning to mitigate intra-class variation. On the other hand, DCMIP also proposes multi-instance proxies based global hard negative sample mining to aid in the discriminative ability of inter-class variance.
ACFL-VIT~\cite{ACFL} uses a ViT model as a feature extractor. First, ACFL-VIT employs discriminative feature learning to extract easy sample features from unlabeled training images and estimates high confidence pseudo labels based on these features. Then, hard sample generation uses the high confidence features to calculate person prototypes, generating hard person features to enhance the discriminative capability of feature learning further. Finally, hard person features and other person prototypes compute contrastive learning, resulting in a discriminative feature learning model.
PASS~\cite{pass} combines ViT and contrastive learning to propose a pre-training method specifically for re-id tasks. PASS introduces [PART] tokens to automatically extract part-level features. It also modifies the contrastive loss to train unsupervised person re-identification models and outperforms the previous best model by 0.7\% and 0.6\% on Market and MSMT17, respectively.

\subsection{Pseudo labels}
\label{sec:related-pseudo}

Due to the lack of labeled data in unsupervised learning tasks, it is common to rely on self-generated pseudo labels to train the model~\cite{eln, ssda, eml_anl}. Pseudo labels can distinguish between positive and negative pairs with greater accuracy. This synergy allows for more effective clustering of similar data points and separation of dissimilar ones, enhancing the discriminative power of the model.
In existing unsupervised clustering methods, only coarse information of global features is often considered, leading to insufficient pseudo labels. PPLR~\cite{pplr} generates PART tokens to enable the model to consider fine-grained clues of regional features during unsupervised clustering, thereby producing robust pseudo labels. The pseudo labels can be refined with the complementary relationship between global and regional features.
SPCL~\cite{spcl} argues that previous pseudo labeling methods suffer from two limitations. The first is that source data is typically only used for pre-training models, wasting valuable true label information in the source data. The second is that when generating pseudo labels, outliers are often discarded. However, these outliers may be valuable samples that are difficult to classify. This paper proposes hybrid memory which utilizes all data for training. The cluster centers of the target domain and the features of the unclustered instances in the target domain can both provide supervision signals for jointly learning discriminative feature representations across two domains.

HPcL~\cite{HPcL} generate hybrid features using the Dynamic Fusion Module and obtain pseudo labels using a cluster algorithm and store them in the memory bank. After that, calculate the contrastive loss using the hybrid features and memory features. Finally, use an orthogonal regularization term to constrain the condition number of the weight Gram matrix, which prevents hybrid features from being concentrated in a compact feature subspace.
MaskPre~\cite{MaskPre} adopt the Dynamic Dropblock Layer to mask region-level features, generating different masked views of a single image. Additionally, a moving-average model and mask prediction head are designed to predict the masked regions and bridge domain gaps across views. MaskPre calculates a pseudo label for one of the moving-average model outputs and computes the cross-entropy loss. Furthermore, leveraging the idea of contrastive learning, it is assumed that different masked views features should have consistent semantics. Accordingly, a consistency loss is proposed to measure the different masked views.
AFC~\cite{AFC} proposes a Group Attention Module to emphasize important features in the input image, utilizing the dependencies between different channels to remove spatial information. Additionally, it learns to generate spatial attention maps by focusing on the pixels that contribute most to network inference, and proposes an Instance Discriminatory Loss based on the concept of contrastive learning. AFC also introduces pseudo labels refinement using Temporal Ensembling and Clustering Consistency for Pseudo Label Propagation to enhance the quality of pseudo labels.
HSP-MFL~\cite{HSP-MFL} designs a Multiple Feature Fusion Module to explore the correlations between multiple semantic and visual features. HSP-MFL then uses multi-head self-attention to generate more discriminative attention-fused features. It combines these features with pseudo labels to form a multi-task contrastive loss for training the model.

\subsection{Prototypes}
\label{sec:related-pseudo}

Prototype-based models have been widely used in recent unsupervised person re-id research~\cite{semi_prototype, CPSPAN, protonce, pipnet}. 
ClusterNCE~\cite{clusternce} points out that previous memory-based contrastive learning works are limited by batch size constraints. Only a small portion of instance features are allowed the update. This leads to inconsistent distributions of small batch oscillations due to rapid updates of the network in each iteration. The algorithm proposed in this paper uses unique prototypes to represent each clustered category and maintains uniqueness throughout the updating process.
IQE~\cite{IQE} introduces a detail enhancement module to improve the feature extraction capability by simultaneously learning from multiple distorted images, thereby reducing pseudo label noise caused by image distortion. IQE method also proposes a low-light enhancement module that improves image clarity by adaptively adjusting pixel values. Both modules work to enhance feature quality, resulting in higher quality pseudo labels. These pseudo labels are then used to obtain cluster prototypes and outlier samples. All these information is used to compute the contrastive loss.
DLCL~\cite{DLCL} proposes Instance-instance Contrastive Learning, which brings original samples and their augmented samples closer through contrastive loss to explore intra-instance similarities. It also introduces Instance-Community Contrastive Learning, which uses a clustering algorithm to group augmented samples, treating each group as a community. DLCL employs contrastive loss to attract similar instances into the same sample community using the community centroid and original samples, capturing inter-instance similarities.

DHCL~\cite{DHCL} uses pseudo labels to obtain cluster samples and outlier samples, storing them in a memory bank and then calculating cluster prototypes. Finally, it employs contrastive loss to explore intra-category similarity within clustered samples and inter-instance discrimination among individual instances.
Han X. \textit{et al.}~\cite{rethinking} posits that sampling strategy is as important as model design and loss function. It addresses deteriorated over-fitting by enhancing statistical stability and, inspired by this, proposes group sampling. First, pseudo labels are used to group samples of the same category. Then, group sampling is performed using information from outlier samples and prototypes, ensuring that each mini-batch contains samples from the same category. Group samples are used for contrastive learning training, reducing the negative impact of individual samples and enhancing intra-category statistical stability.
RTMem~\cite{RTMem} calculates cluster prototypes using pseudo labels and stores features in memory. Unlike the general momentum update strategy, RTMem updates the cluster prototype memory with randomly sampled instance features from the current mini-batch. It then creates a sample-to-instance contrastive loss using instances of the same category as positive samples and all instances as negative samples. Additionally, RTMem constructs a sample-to-cluster contrastive loss using prototypes to enhance feature representation capability.
ISE~\cite{ise} uses prototypes generated from data to produce support samples near the boundaries of clusters. Support samples are computed with real features to calculate loss, aiming to make the original data and the support samples close to the centroids of their respective clusters.
CCIOL~\cite{CCIOL} uses the sample online inter-camera K-reciprocal nearest neighbors to soften prototype pseudo labels and generate instance-level multi-labels, addressing the negative impact of camera distribution differences. Additionally, CCIOL employs contrastive learning loss at two levels to correct erroneous proximities between samples, promoting inter-class and intra-class separation.

The above methods propose excellent systems to address the unsupervised person re-identification task, but some issues remain.
Previous methods using ViT do not consider the issue of ViT patch noises~\cite{pass, ACFL}. 
Many current methods utilize instance and cluster level information, but they often ignore the potential value of outlier samples~\cite{CCIOL, ACFL} or require additional model costs to improve performance~\cite{CCIOL}.
In this paper, TCMM mitigates the ViT patch noise issue, enhancing the quality of features extracted by ViT. Without increasing model cost or altering the pseudo label process, TCMM considers both prototype and instance-level information. It further explores the potential value of outliers and encourages the model to learn from difficult samples, enabling it to cover various data variations. When selecting negative samples, consider the overall distribution of the negative candidates to reduce the impact of the uncertainty caused by random sampling on the model.

% ===========================  Method  ===========================%
\section{Method}
\label{sec:method}

\begin{figure}[H]
\centering
\includegraphics[width=\linewidth]{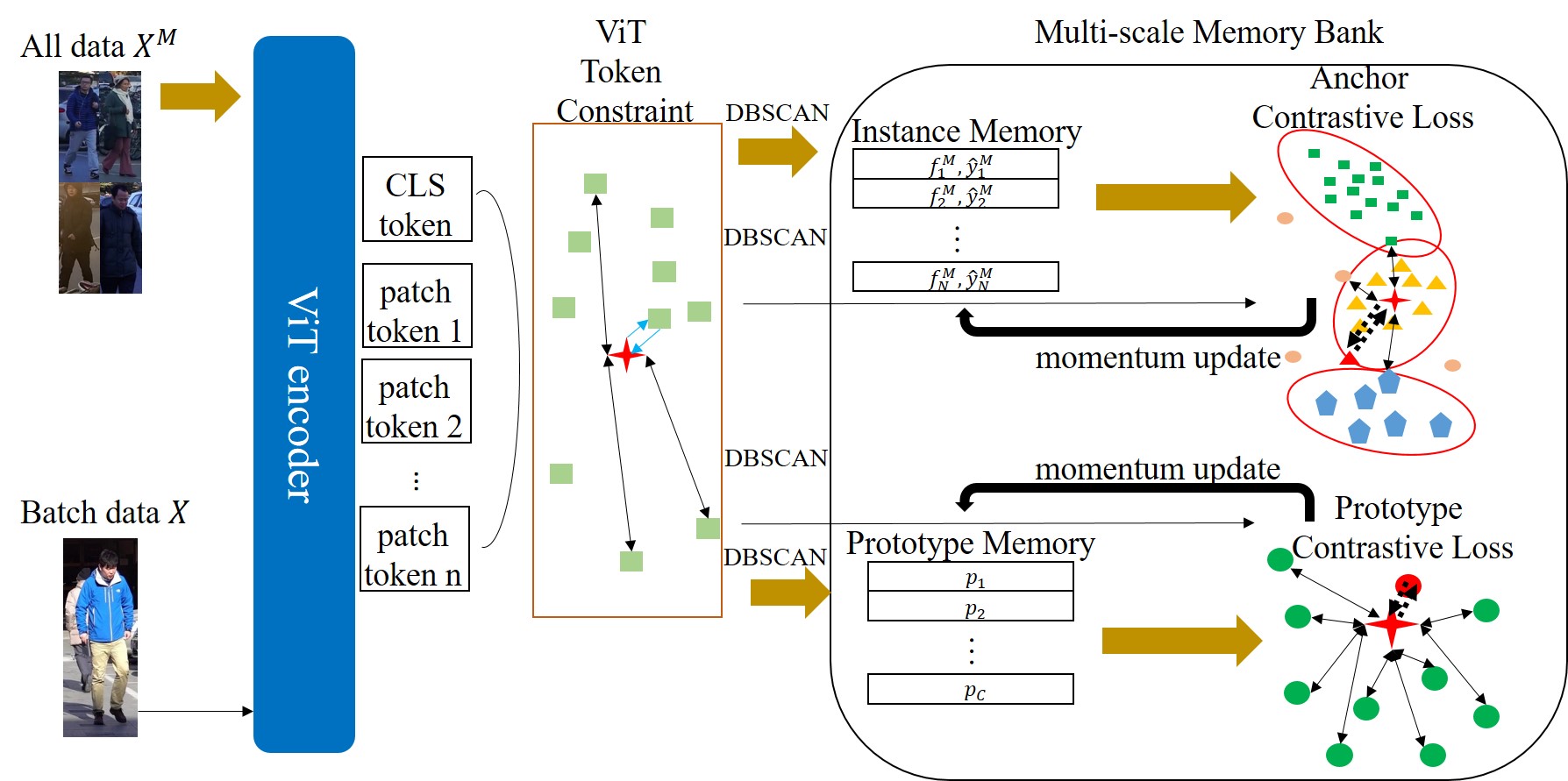}
\caption[The overall architecture of TCMM.]{The overall architecture of TCMM. A ViT encoder processes the input data to obtain token features. All token features are input into a transformer token constraint process to mitigate the negative impact of patch noises. Features of all training data are assigned pseudo labels using clustering algorithms (DBSCAN is used as an example here). All data is adopted to form instance and prototype level memory using pseudo labels. Then anchor contrastive loss and prototype contrastive loss are computed with batch input and memory. Both memory modules are updated using the momentum update technique after contrastive learning.} 
\label{fig:overall}
\end{figure}

% problem formulation
Given person re-id training set $X^M= \{x_n^M\}_{n=1}^N$ with $N$ images and batch data $X= \{x_b\}_{b=1}^B$ with $B$ samples. All samples are encoded to instance features $f_n^M, f_b$ by ViT encoder $E_{\theta}$ and assigned pseudo labels $\hat{y}_n^M, \hat{y}_b$ by cluster method.
The proposed TCMM aims to strengthen the encoder $E_{\theta}$ to learn more discriminative features, enhancing the model robustness and diversity.  
TCMM limits the interference of patch noise through the ViT token constraint, and the multi-scale memory bank improves the accuracy of pseudo labels. The system ultimately enhances the performance of unsupervised person re-id.
The overall architecture of ViT Token Constraint and Multi-scale Memory bank (TCMM) is depicted in Figure~\ref{fig:overall}. Section \ref{sec:method-transformer} introduces how to restrict ViT tokens to reduce the negative impact of patch noise, while section \ref{sec:method-memory} details how to utilize multi-scale memory bank to enable $E_{\theta}$ to learn more discriminative features.

% ===========================  Method: Transformer Token Constraint  ===========================%
\subsection{ViT Token Constraint}
\label{sec:method-transformer}
The success of self-attention heavily relies on the correlation between patch tokens. However, not all patches in the patch embedding process are helpful. These blocks may introduce patch noises when encountering occlusions, backgrounds, or non-target pedestrians. 
% 8/6
These noises could cause irrelevant regions to be perceived as similar to the target pedestrian areas, leading to attention being dispersed across different regions. This dispersion of attention might result in the model focusing on irrelevant regions when computing feature representations. Moreover, when running into non-target pedestrians, some weights might erroneously be assigned to these interfering regions, causing even more harm to the model and affecting its accuracy in identifying individuals.
Therefore, this paper proposes a ViT token constraint function to limit the patch token features obtained from the encoder, thereby reducing the impact of patch noise on the ViT.

Since during training, it is not possible to know which patch token features are generated by noise, ViT token constraint utilizes the concept of contrastive learning. This design aims to push those potential patch noise features away from the image feature as much as possible. Typically, contrastive learning treats all other samples as negative samples. However, based on the observations of person re-id data, only a few patches are usually noises. Treating the majority of patch token features as negative samples is not appropriate. Therefore, the ViT token constraint first defines a patch hyper parameter $R$ to control the number of negative samples regarding the issue of patch noise features. Equation (\ref{equ:patch_rate}) is the description of $R$ design.
\begin{equation}
\label{equ:patch_rate}
\begin{aligned}
    R=\left \lfloor I*\alpha  \right \rfloor 
\end{aligned}
\end{equation}
where $I$ is the patch number and $\alpha$ means the rate to determine the usage of negative patch token features.

\begin{figure}[htbp]
\centering
\includegraphics[width=\linewidth]{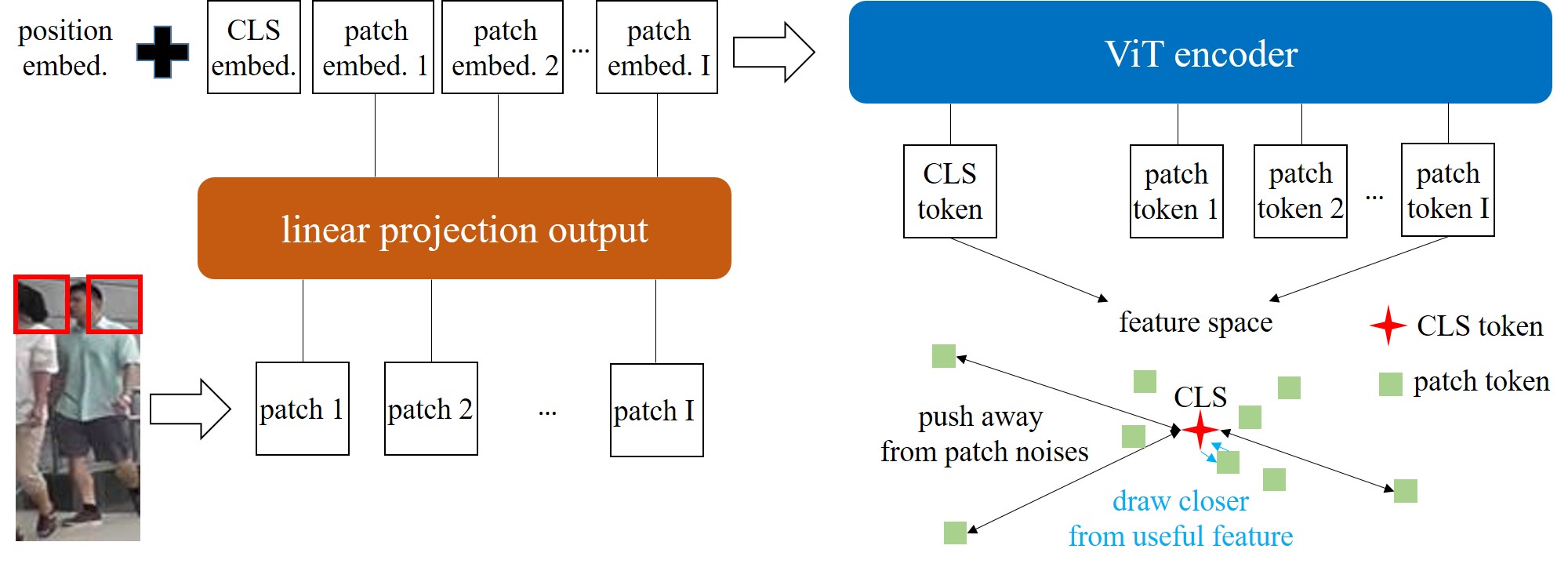}
\caption[The process of ViT token constraint.]{The process of ViT token constraint.The output CLS feature $f_b$ of the encoder is compared for similarity with all patch token features. Drawing from the concept of contrastive learning, $f_b$ is brought closer to the token with the highest similarity and pushed further away from $R$ tokens with the lowest similarity.}
\label{fig:contraint_process}
\end{figure}

To constrain the ViT token features, this paper draws inspiration from the concept of contrastive learning. The system computes the similarity between the image feature $f_b$ and all patch token features $f_i^t$. Then, the method selects the patch token feature with the highest similarity as the positive sample and chooses $R$ least similar patch token features as negative samples. ViT token features encourage the ViT to rely more on useful patch information and push away from $f_b$ and patch noises, reducing the ViT encoder reliance on noises. This limits the negative impact of patch noises on the ViT. The overall ViT token constraint concept is illustrated in Figure~\ref{fig:contraint_process}. Equation~(\ref{equ:constraint}) is the ViT token constraint function.

\begin{equation}
\label{equ:constraint}
\begin{aligned}
    L_{constraint}=-log\frac{exp(f_b\cdot f_+^t/\tau)}{exp(f_b\cdot f_+^t/\tau)+\sum_{r=1}^{R}exp(f_b\cdot f_r^t/\tau)}
\end{aligned}
\end{equation}
where $f_b$ indicates input batch instance features, $f_+^t$ means the positive patch token feature and $f_r^t$ is the $r$-th negative patch token feature. $\tau$ is the temperature hyper-parameter to control the smoothness of the probability distribution. 
Using the sum of positive and negative samples in the denominator allows the term after $-log$ to approach the range 0 to 1. In minimizing loss, it is necessary to make the inside of $-log$ close to 1. The positive sample pairs must be much greater than the negative sample pairs, meaning the similarity with the positive samples must be much greater than with the negative samples. The exponential function performs nonlinear amplification, which helps the model more strongly distinguish similar samples from different samples.
This formula aims to impose constraints on the ViT encoder, ensuring that the model output feature $f_b$ is as far away from patch noises as possible. 

% ===========================  Method: Multi-Scale Memory Loss  ===========================%
\subsection{Multi-Scale Memory Bank}\label{sec:method-memory}
In the model training stage, the feature update of instances in an iteration might not be consistent with the features of the same cluster because of data imbalance limitations. 
Moreover, the common clustering methods may result in outlier samples that are not assigned to any cluster. For example, the commonly used DBSCAN algorithm uses -1 as a pseudo label for outlier samples that cannot be clustered.
Some existing methods choose to discard outlier samples that are not assigned to any cluster. These methods have demonstrated that discarding outlier samples strategy can improve model performance. Since re-id models are often sensitive to noise, not using unlabeled outlier samples can reduce the negative impact of potential noise on the model. However, outlier values may represent valuable yet difficult-to-classify samples. Additionally, many outliers are generated early in training. Discarding these outliers may harm the final performance and limit the diversity of the model. Therefore, this paper aims to adopt the architecture of the memory bank and the strategy of discarding outlier samples while also exploring the latent value of outlier samples. By reducing the negative impact of potential noise on the model and considering the potential value of outlier samples, the approach seeks to strike a balance between improving performance and preserving diversity.
Section~\ref{sec:method-proto} describes the proposed prototype level memory, which adopts a strategy of discarding outlier samples. The remaining samples are clustered to obtain prototypes, which are then used to construct a prototype memory. Furthermore, this paper introduces the Prototype Contrast Loss function to address instance level and cluster level feature inconsistencies caused by batch size. The prototype contrast Loss function aims to minimize the discrepancy between feature representations at the instance level and cluster level, thereby improving the robustness of the model.
Section~\ref{sec:method-anchor} introduces an instance level memory based on Memory Bank~\cite{memory_bank}. It utilizes all data containing outlier samples to construct an instance memory bank. Additionally, the instance level memory proposes the Anchor Contrastive Loss function to encourage the model to consider more challenging samples and explore the value of outlier samples, thus enhancing the model generalization and performance. 

% ===========================  Method: Prototype Contrast Loss  ===========================%
\subsubsection{Prototype Memory and Prototype Contrast Loss}\label{sec:method-proto}

\begin{figure}[H]
\centering
\includegraphics[width=\linewidth]{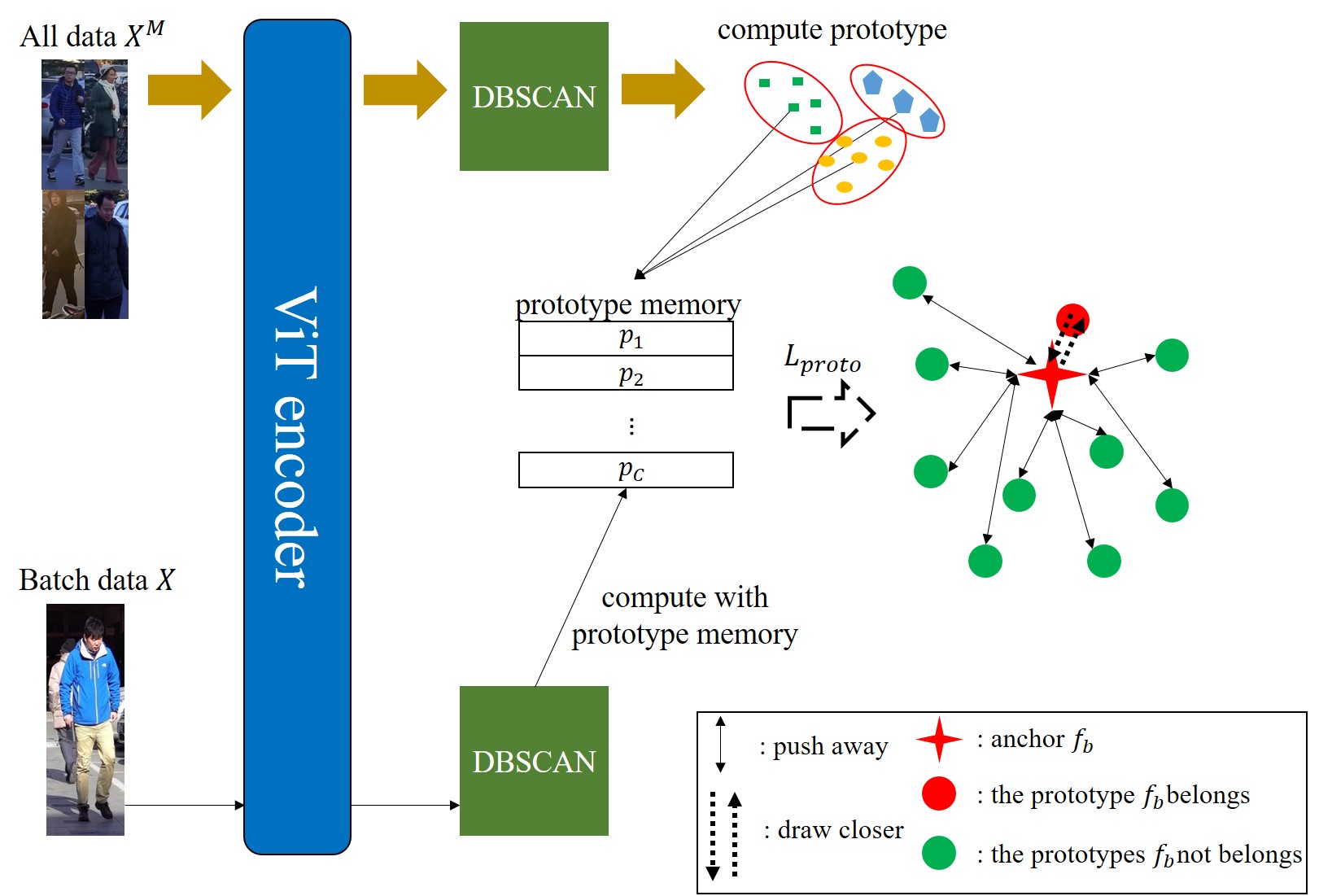}
\caption[The process of prototype contrastive loss.]{The process of prototype contrastive loss with DBSCAN example. The input data $X^M, X$ are fed into the ViT encoder to extract features, which are then input into DBSCAN to obtain pseudo labels and compute prototypes. These prototypes are stored in the prototype level memory and compared with the anchor $f_b$ to calculate the prototype contrastive learning loss function.} 
\label{fig:prototype_loss}
\end{figure}

After applying a clustering algorithm, the training data $X^M$ without outlier samples is divided into $C$ clusters and stored in the prototype level memory. Prototype contrast loss utilizes the prototypes of the cluster as positive and negative samples to ensure feature consistency. The model consistently considers representative features of each cluster during training by treating the centroid of each class cluster as a positive sample. This ensures the stability and consistency of the model feature representation. Using the cluster centroid from other clusters as negative samples helps the model learn to distinguish features between different clusters.
The process of prototype memory and prototype contrast loss is illustrated in Figure~\ref{fig:prototype_loss}. This design can force the model to learn to distinguish the feature representations of different clusters, thereby improving the model classification ability. Equation~(\ref{equ:pro_ini}) is the prototypes design.

\begin{equation}
\label{equ:pro_ini}
\begin{aligned}
    p_c=\frac{\sum _{{f_j^M}\in \mathcal{O}_c} f_j^M}{|\mathcal{O}_c|} 
\end{aligned}
\end{equation}
where $\mathcal{O}_c$ indicates the $c$-th cluster set, $f_j^M$ means the $j$-th instance feature in $\mathcal{O}_c$, $|\mathcal{O}_c|$ is the sample number of $c$-th cluster set and $p_c$ indicates prototype of the $c$-th cluster. The centroid feature of each cluster $\mathcal{O}_c$ is regarded as the cluster prototype $p_c$ in this equation. The prototypes $p_c$ obtained from this equation are also stored in a prototype level memory $M^{proto} = \{ p_1, p_2,...,p_C\}$. 

Prototypes ensure that the features of each cluster receive sufficient consideration since each cluster prototype is computed from all samples in that cluster. This avoids the impact of a small number of samples in certain clusters on the training effectiveness. After obtaining prototypes, the InfoNCE loss~\cite{infonce} can be rewritten as the prototype contrast loss. Equation (\ref{equ:prototype}) is the rewritten loss.

\begin{equation}
\label{equ:prototype}
\begin{aligned}
    L_{proto}=-log\frac{exp(f_b\cdot p_+/\tau)}{\sum_{c=1}^{C}exp(f_b\cdot p_c/\tau)}
\end{aligned}
\end{equation}
where $p_+$ represents the prototype of the same class, $p_c$ indicates the $c$-th prototype, and $\tau$ is the temperature hyper-parameter. Batch input data is pulled closer to the prototype of their respective clusters while being pushed away from prototypes of other clusters. Overall, the design of the prototype contrast loss considers both the consistency and stability of features, the discrimination between classes, and the issue of data imbalance, thereby offering certain advantages.

% ===========================  Method: Anchor Contrastive Loss  ===========================%
\subsubsection{Instance Memory and Anchor Contrastive Loss}\label{sec:method-anchor}
As mentioned earlier, previous methods have demonstrated that discarding outliers can reduce the negative impact of potential noise on the model. Outlier values may represent valuable yet difficult-to-classify samples. Discarding these outliers may harm the final performance and limit the diversity of the model.
To fully consider the value of all samples, the instance level memory method first extracts features from the training set $X^M$ and stores them in the memory bank $M^{ins}= \{f_1^M, f_2^M,...,f_N^M\}$. A clustering algorithm such as DBSCAN~\cite{dbscan} is then used to obtain pseudo labels $\hat{Y}^M$ for all features. During training, the batch features $f_b$ and pseudo labels $\hat{y}_b$ of input batch data $x$ are also obtained through the same process. To encourage the model to consider more challenging samples and improve its generalization, this paper regards $f_b$ as an anchor and rewrites the InfoNCE loss as the proposed anchor contrastive loss. Firstly, the positive samples are replaced by the least similar samples within the same class as the input sample in $M^{ins}$, making the model more challenging to distinguish between different instances within the same class. The positive sample design helps the model learn more discriminative features and improves its generalization ability.

In a typical contrastive learning loss function, negative samples are usually set to be all samples other than the positive sample. However, in the design of the anchor contrastive loss, the intention is for the model to consider more challenging samples. The model may consider dissimilar samples from different clusters and hinder the learning process if the number of negative samples $\kappa$ is too large. On the other hand, when the number of negative samples $\kappa > 1$, the effect tends to be better as it increases the diversity of negative samples and helps the model better learn to distinguish features between different categories. In summary, an appropriate number of negative samples $\kappa$ can increase the diversity of negative samples, aiding the model in learning feature representations.

Accordingly, the negative samples are replaced by the $\kappa$ most similar samples from different cluster to the input sample in $M^{ins}$. Setting negative samples as the $\kappa$ most similar samples from different categories forces the model to learn better to distinguish features between different categories. The negative sample setting helps increase the diversity between training samples, ensuring that the model has a better perception of differences between various categories during the learning process, thereby improving the model robustness and classification ability. Equation~(\ref{equ:anchor}) is the overall design of the anchor contrastive loss.

\begin{equation}
\label{equ:anchor}
\begin{aligned}
    L_{anchor}=-log\frac{exp(f_b\cdot f_+^{ins}/\tau)}{exp(f_b\cdot f_+^{ins}/\tau)+\sum_{k=1}^{\kappa}exp(f_b\cdot f_k^{ins}/\tau)}
\end{aligned}
\end{equation}
where $f_+^{ins}$ indicates the least similar instance feature within the same class in $M^{ins}$, $\kappa$ is the number of negative samples, $f_k^{ins}$ means $\kappa$ most similar instance from different class in $M^{ins}$ and $\tau$ is the temperature hyper-parameter to control the smoothness of the probability distribution. The anchor contrastive loss $L_{anchor}$ selects the hardest-to-distinguish sample from the same cluster as the input batch data $f_b$ as the positive sample, then uses $\kappa$ samples from other clusters that are easily confused as negative samples. $L_{anchor}$ can focus more on learning from difficult samples through this contrastive learning sample selection strategy.

\begin{figure}[H]
\centering
\includegraphics[width=\linewidth]{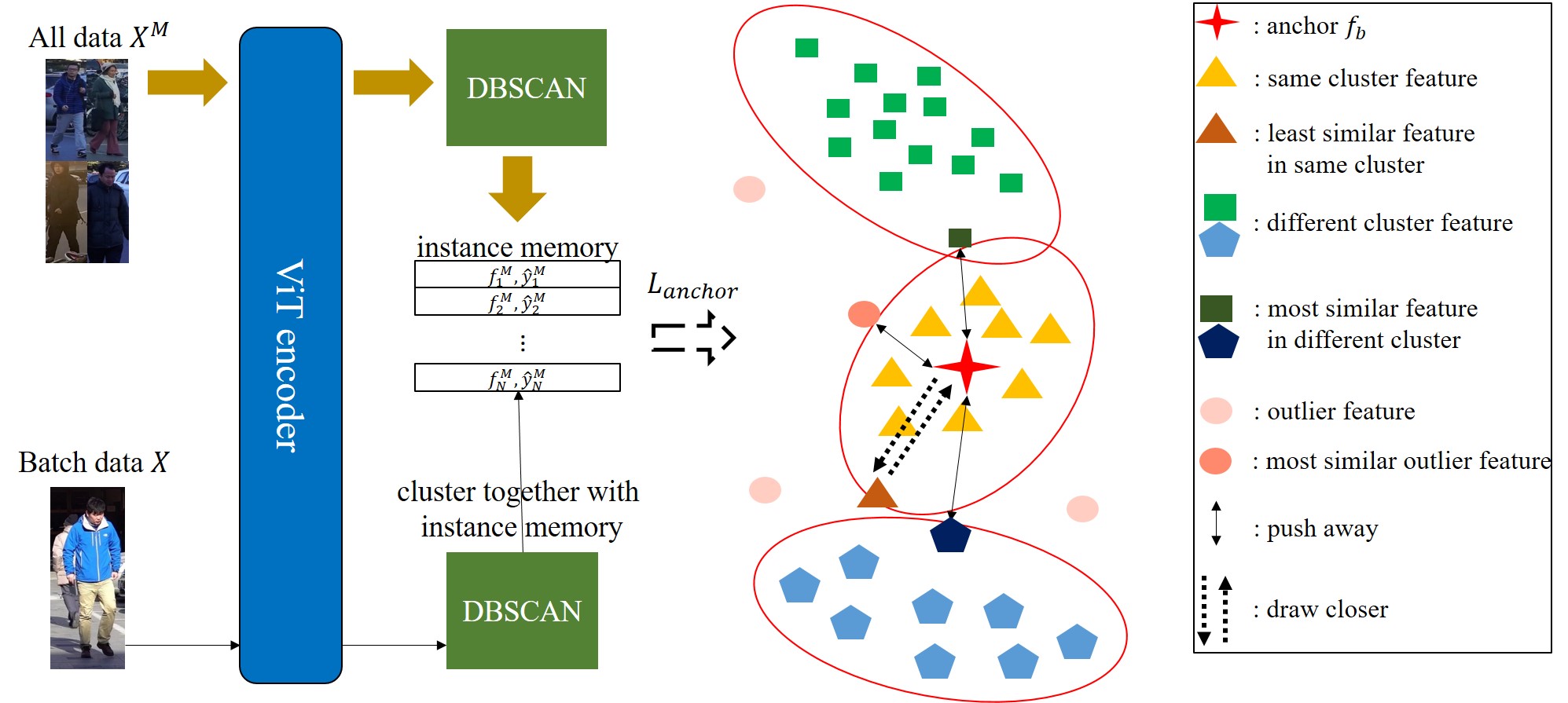}
\caption[The process of anchor contrastive loss.]{The process of anchor contrastive loss with DBSCAN example. The input data $X^M, X$ are fed into the ViT encoder to extract features, which are then input into DBSCAN to obtain pseudo labels. This information is stored in the instance level memory and compared with the anchor $f_b$ to calculate the contrastive learning loss function.}
\label{fig:anchor_loss}
\end{figure}

$L_{anchor}$ encourages the model to learn from difficult samples through different sample selection strategy. This novel contrastive loss function is more reasonable in sample selection and more challenging and diverse. This design helps improve the model generalization ability and robustness and better utilizes all training data. Therefore, better results can be expected in practical applications. 
Notice that $L_{anchor}$ considers all data, including outlier samples. Unlike conventional methods that ignore outlier samples, negative samples in $L_{anchor}$ may include outlier samples. Through this design, it is possible to discover whether outliers have potential value. The model can better utilize the information in the training data. $L_{anchor}$ encourages learning from difficult samples and leverages outlier samples that may contain useful information, thereby improving training efficiency and effectiveness.
Overall, this new contrastive loss function features a more reasonable sample selection process, making it more challenging and diverse. This enhances the model generalization ability and robustness while leveraging the training data better. The overall instance memory and anchor contrastive loss process is illustrated in Figure~\ref{fig:anchor_loss}. 

% anchor contrastive loss discussion
\subsubsection{Momentum Update}\label{sec:method-momentum}
% momentum update 的理由與流程
Encoder $E_{\theta}$ updates during every training iteration, while the features in memory are generated using an older version of $E_{\theta}$. This discrepancy can lead to memory and batch features not being represented by a similar $E_{\theta}$. Inspired by previous works such as \cite{moco, spcl, mean_teacher}, momentum update is employed for the two memories. 
% prototype memory update 介紹
After computing the prototype contrast loss, a momentum update strategy is employed to update the prototype memory. Equation~(\ref{equ:proto_update}) is the definition of prototype momentum update.

\begin{equation}
\label{equ:proto_update}
\begin{aligned}
    \forall p_c \in M^{proto}, p_c \gets \mu p_c + (1-\mu)f_b
\end{aligned}
\end{equation}
where $\mu$ is the momentum coefficient. When $\mu$ is 1, the features in the prototype memory remain unchanged. Whereas when $\mu$ is 0, the features in the prototype memory are exactly the same as the input batch features $f_b$.
This equation blends the original prototype features and the batch instance features in proportion to $\mu$ and then updates the prototype features $p_c$ in $M^{proto}$. 
$M^{proto}$ is updated by batch input in each iteration. Compared to recalculating the prototype after training with all the data, this approach keeps the prototype up-to-date and is also faster.

On the other hand, the instance level memory is also updated by momentum update strategy. Equation~(\ref{equ:ins_update}) is the momentum update process of instance level memory.

\begin{equation}
\label{equ:ins_update}
\begin{aligned}
    \forall f_n^M \in M^{ins}, f_n^M \gets \mu f_n^M + (1-\mu)f_b
\end{aligned}
\end{equation}
where $\mu$ is the momentum coefficient. When $\mu$ is 1, the features in the instance memory remain unchanged. Whereas when $\mu$ is 0, the features in the instance memory are exactly the same as the input batch instance features $f_b$. The feature inconsistencies caused by rapidly changing encoder can be avoided by employing this updating strategy.
%%%%%%%%%%%%%%%       alg  %%%%%%%%%%%%%%%%%%%%%%%%%%%%%
\begin{algorithm}
\caption{TCMM algorithm with DBSCAN clustering}
\label{alg:TCMM}
\begin{algorithmic}[1]
\Require ViT encoder $E_{\theta}$, training epoch $T$, batch number $B$, negative patch token rate $\alpha $, temperature hyper-parameter $\tau$, momentum coefficient $\mu$
\Input Unlabeled training data $X^M = \{x_n^M\}_{n=1}^N$, Input batch data $X = \{x_b\}_{b=1}^B$
\State \textbf{Init.}: All data features $F^M = \{f_n^M\}_{n=1}^N = E_{\theta}(X^M)$
\For{epoch 1 to $T$}
    \State All data pseudo labels $\hat{Y}^M = \{\hat{y}_n^M\}_{n=1}^N$ = DBSCAN($F^M$)
    \State Instance memory $M^{ins} = \{f_n^M, \hat{y}_n^M\}_{n=1}^N$
    \State Calculate $P=\{p_c\}_{c=1}^C$ according to Eq. \ref{equ:pro_ini}
    \State Prototype memory $M^{proto}=P$
    \For{iteration 1 to $iter\_num$}
        \State Sample $B$ data $X = \{x_b\}_{b=1}^B$ from $X^M$
        \State Sample $B$ pseudo labels $\hat{Y} = \{\hat{y}_b\}_{b=1}^B$ corresponding to $X$ from $\hat{Y}^M$
        \State Batch features $F=\{f_b\}_{b=1}^B = E_{\theta}(X)$
        \State Calculate $L_{constraint}$ with each batch feature $f_b$ according to Eq. \ref{equ:constraint}
        \State Calculate $L_{proto}$ with $f_b, p_c$ according to Eq. \ref{equ:prototype}
        \State Calculate $L_{anchor}$ with $f_b, f_n^M$ according to Eq. \ref{equ:anchor}
        \State Update $M^{proto}$ with $f_b, p_c$ according to Eq. \ref{equ:proto_update}
        \State Update $M^{ins}$ with $f_b, f_n^M$ according to Eq. \ref{equ:ins_update}
        \State Optimize $E_{\theta}$ by minimizing ($L_{constraint}+L_{proto}+L_{anchor}$)
    \EndFor
\EndFor
\Output Optimized $E_{\theta}$
\end{algorithmic}
\end{algorithm}

In summary, with the ViT token constraint module, the ViT model can better mitigate the adverse effects caused by patch noises. 
Furthermore, prototype memory and prototype contrast loss ensure feature consistency and stability. Meanwhile, by employing instance memory and anchor contrastive loss, the model can better explore the hidden value of outlier data and facilitate learning more discriminative features. The process of TCMM is described in Algorithm \ref{alg:TCMM}.
Equation~(\ref{equ:total}) is the overall TCMM objective loss function.

\begin{equation}
\label{equ:total}
\begin{aligned}
    L_{total} = \lambda_{con}L_{constraint}+\lambda_{pro}L_{proto}+\lambda_{an}L_{anchor}
\end{aligned}
\end{equation}
where $\lambda_{con},~\lambda_{pro},~\lambda_{an}$ are exploited for balancing each term in the loss function.

% ===========================  Experimental Results  ===========================%
\section{Experimental Results}
\subsection{Dataset and Implementation}
\label{sec:dataset}
In this paper, we use two large-scale person re-ID dataset: Market-1501~\cite{market1501} and MSMT17 \cite{msmt17} for our model evaluation. We don't adopt DukeMTMC-reID dataset because it has been taken down.
Standard evaluation metrics like mean Average Precision (mAP) and Rank-1 accuracy from the cumulative matching curve (CMC)~\cite{market1501} are utilized in the experiments on both datasets.

We construct our ViT model according to~\cite{dino,pass}.
Then, we conduct the pre-training process using~\cite{lup,PUL}. We use the ViT encoder along with general contrastive loss~\cite{moco} as our baseline. We first resize the input images to $256 \times 128$. Then the input data is divided into $I$ patches. We adopt a patch size of $16 \times 16$. Therefore, the number of patches $I=128$. The PASS~\cite{pass} process utilizes a [CLS] token and $Z$ additional [PART] tokens for training. Following the PASS process, we set $Z$ to 3 and design image feature $f$ as follows.

\begin{equation}
\label{equ:cls}
\begin{aligned}
    f =[CLS] \oplus  \frac{ {\textstyle \sum_{z=1}^{Z}} [PART]_z}{Z} 
\end{aligned}
\end{equation}

The $\oplus$ means concatenating and the output dimension of $f$ using this process is 768. Our approach trains with a batch size of 512 for 80 epochs using the SGD optimizer. The learning rate is set to $3.5e-4$ and multiplied by 0.1 every 20 epochs. We set $\tau$ and $\mu$ to 0.05 and 0.2 following previous works~\cite{spcl,pass}. We set $\lambda_{con},~\lambda_{pro},~\lambda_{an}$ to 1 by empirical. All experiments adopt DBSCAN as the clustering algorithm. We discuss the settings of $\alpha$ in $L_{constraint}$ and $\kappa$ in $L_{anchor}$ in the following section~\ref{sec:ablation}.
Our architecture is implemented in Pytorch with GTX 1080 Ti GPUs. 

\subsection{Ablation Studies}
\label{sec:ablation}

\begin{figure}[t]
\centering
\includegraphics[width=0.7\linewidth]{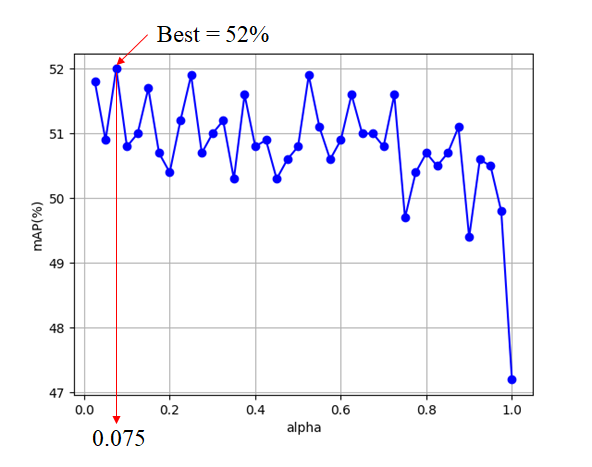}
\caption[Study of $\alpha$ in $L_{constraint}$.]{Study of $\alpha$ in $L_{constraint}$ on MSMT17 dataset. This experiment evaluates the impact of $\alpha$ on model performance with fixed settings $L_{proto}$ and $L_{anchor}$. When $\alpha$ is 0.075, the best mAP performance of 52\% is achieved.}
\label{fig:alpha}
\end{figure}

\textbf{$\alpha$ in ViT Constraint.}
We conduct an ablation study on MSMT17 to investigate the impact of different $\alpha$ setting in $L_{constraint}$. 
We investigate the impact of $\alpha$ on model performance with fixed settings $L_{proto}$ and $L_{anchor}$. We test several settings and calculate mAP: from 0.025 to 1, with an interval of 0.025. 
The experiments in Figure~\ref{fig:alpha} show that the best mAP performance is achieved when $\alpha$ is set to 0.075 with 52\%, respectively. We can observe from Figure~\ref{fig:alpha} that the mAP performance shows a decreasing trend as $\alpha$ increases. The oscillation of mAP performance between $\alpha$ = 0.025 to 0.6 is likely due to normal perturbations during training.
This experiment confirms our hypothesis. Since we cannot determine how many patches are noise, the patch noise in a single image is usually not excessive based on our observation. 
Therefore, treating all patches as negative samples in traditional contrastive learning is inappropriate. Using a small number of patch tokens as negative samples is more beneficial for the model. 

\textbf{$\kappa$ in Anchor Loss.}
We conduct an ablation study on MSMT17 to investigate the impact of different $\kappa$ setting in $L_{anchor}$. 
We use fixed settings prototype loss $L_{proto}$ and ViT constraint $L_{constraint}$ to validate the impact of $\kappa$ in $L_{anchor}$ on performance. We test settings of $\kappa$ and calculate mAP and Rank-1. Experiment results in Table~\ref{tab:kappa} demonstrate that our anchor loss $L_{anchor}$ design is effective.

When $\kappa$=4, we achieve the best mAP/Rank-1 performance at $52.0\%/78.4\%$. Compared to $\kappa=1$, setting negative samples to multiple samples most dissimilar to the input sample from different classes increases the training difficulty, forcing the model to learn more discriminative feature representations. The mAP performance shows a decreasing trend as $\kappa$ increases. When $\kappa$ is too large, the model may focus too much on dissimilar samples from other clusters. These samples provide little assistance and may even increase the complexity of training, leading to convergence difficulties or easy overfitting of the training data.

\begin{table}[h]
\setlength{\belowcaptionskip}{12pt}
\center
\caption[Study of $\kappa$ in $L_{anchor}$.]{Study of $\kappa$ in $L_{anchor}$ on MSMT17. This experiment uses fixed settings prototype loss $L_{proto}$ and ViT constraint $L_{constraint}$ to examine the impact of $\kappa$ on model performance. Bold indicates the best performance.}
\begin{tabular}{ccc}
\hline 
$\kappa$ &    mAP(\%) & Rank-1(\%) \\ \hline
1 & 50.9 & 78.1  \\
2 & 50.8 & 78.0 \\
3 & 51.5 & 78.1  \\
4 & \textbf{52.0} & \textbf{78.4} \\
5 & 50.5 & 77.5 \\
6 & 50.7 & 77.4 \\
7 & 50.8 & 77.6 \\
8 & 51.3 & 78.1 \\
16 & 50.8 & 77.5 \\
32 & 50.9 & 77.4 \\
64 & 48.2 & 75.1 \\ \hline
\end{tabular}
\label{tab:kappa}
\end{table} 

\textbf{Outlier sample strategy in Anchor Loss.}
We conduct an ablation study on the MSMT17 dataset to investigate the impact of different outlier sample strategies in $L_{anchor}$.
We validate the effect of the outlier sample strategy on the model with $L_{constraint}$ ($\alpha=0.075$) and $L_{anchor}$ ($\kappa =4$).
We test two settings, with and without outlier samples, to compute mAP and Rank-1. From Table~\ref{tab:outliers}, the experiment demonstrates that our outlier sample strategy idea is effective. When $L_{anchor}$ includes outlier samples, the mAP/Rank-1 performance exceeds that without outlier samples by $1.7\%/1.3\%$. Not using all outlier samples may lead to the model lacking a global understanding of the entire dataset, resulting in poor performance when dealing with unknown or difficult-to-identify samples. This can cause a decrease in the model ability to identify certain classes or samples.

\begin{table}[h]
\setlength{\belowcaptionskip}{12pt}
\center
\caption[Study of outlier sample strategy in $L_{anchor}$.]{Study of outlier sample strategy in $L_{anchor}$ on MSMT17. This experiment uses $L_{constraint} + L_{proto} + L_{anchor}$ to test the impact of the outlier sample strategy on model performance. "w/" denotes using all outlier samples in $L_{anchor}$, while "w/o" indicates ignoring all outlier samples in $L_{anchor}$. Bold indicates the best performance.} 
\begin{tabular}{ccc}
\hline 
Outliers &    mAP(\%) & Rank-1(\%) \\ \hline
w/o & 50.3 & 77.1  \\
w/ & \textbf{52.0} & \textbf{78.4} \\ \hline
\end{tabular}
\label{tab:outliers}
\end{table}

\begin{table}[h]
\setlength{\belowcaptionskip}{12pt}
\center
\caption[Modules Cross-Validation.]{The modules cross-validation on MSMT17 dataset. Baseline means using the ViT encoder with general contrastive loss. 'Baseline + $L_{constraint}$' is training with the $L_{constraint}$. '$L_{constraint}$+$L_{proto}$' represents using $L_{proto}$ to replace the contrastive loss in baseline. $L_{constraint}$ + $L_{proto}$ + $L_{anchor}$ indicates using all the loss functions. Bold indicates the best performance.} 
\resizebox{0.7\textwidth}{!}{
\begin{tabular}{ccc}
\hline 
Modules &    mAP(\%) & Rank-1(\%) \\ \hline
Baseline & 25.8 & 54.6  \\
Baseline+$L_{constraint}$ & 37.5 & 62.3  \\
$L_{constraint}$ + $L_{proto}$ & 42.0 & 67.3 \\
$L_{constraint}$ + $L_{proto}$ + $L_{anchor}$ & \textbf{52.0}&\textbf{78.4} \\ \hline
\end{tabular}}
\label{tab:cross}
\end{table}

\textbf{Modules Cross-Validation.}
We conduct an ablation study on the MSMT17 dataset to investigate the impact of different modules. We use $L_{constraint}$ with $\alpha=0.075$ and $L_{anchor}$ with $\kappa=4$ to evaluate the effectiveness of different modules. We test configurations including baseline, baseline + $L_{constraint}$, $L_{constraint}$ + $L_{proto}$, and $L_{constraint}$ + $L_{proto}$ + $L_{anchor}$ to calculate mAP/Rank-1. The experiment in Table~\ref{tab:cross} demonstrates that each module effectively improves the model performance. When using the baseline + $L_{constraint}$ configuration, the mAP/Rank-1 performance exceeds the baseline by $11.7\%/7.7\%$. With the $L_{constraint}$ + $L_{proto}$ configuration, the mAP/Rank-1 performance surpasses the baseline + $L_{constraint}$ configuration by $4.5\%/5.0\%$. Finally, the mAP/Rank-1 performance can improve on the above settings by $10.0\%/11.1\%$ using all the loss functions. 
\subsection{Comparison with Unsupervised Person Re-id Methods}
\label{sec:comparison}

We compare our TCMM performance with existing state-of-the-art methods on unsupervised person re-id in Table~\ref{tab:comparison}. The experimental results show that our method significantly outperforms previous unsupervised person re-id works. Our TCMM architecture achieves 90.5\%/96.0\% mAP and Rank-1 on the Market1501 dataset. On the other hand, our TCMM architecture achieves 52.0\%/78.4\% mAP and Rank-1 on the MSMT17 dataset. This surpasses the state-of-the-art results by 2\%/1.1\% mAP/Rank-1 and 6.8\%/3.3\% mAP/Rank-1, respectively. It can be noted that our TCMM not only achieves excellent results on the Market1501 dataset but also achieves outstanding performance on the more challenging MSMT17 dataset. This indicates that our TCMM model not only provides excellent recognition performance but also offers remarkable robustness. 

\begin{table}[t]
\setlength{\belowcaptionskip}{12pt}
\center
\caption[Comparison with Unsupervised Re-id Methods.]{The comparison with state-of-the-art unsupervised person re-id methods on Market1501 and MSMT17 dataset. Bold indicates the best performance, while underline represents the second best performance.}
\resizebox{0.9\textwidth}{!}{
\begin{tabular}{ccc}
\hline 
Method &  \makecell[c]{Market1501 \\ mAP(\%)/Rank-1(\%)} & \makecell[c]{MSMT17 \\ mAP(\%)/Rank-1(\%)} \\ \hline
MaskPre~\cite{MaskPre} (PR'22)  & 77.7/90.4 & 26.7/58.5 \\
X. Han et al.~\cite{rethinking} (TIP'23)  & 79.2/92.3 & 24.6/56.2 \\
MCRN~\cite{mcrn} (AAAI'22)& 80.8/92.5 & 31.2/63.6 \\
SECRET~\cite{secret} (AAAI'22)& 81.0/92.6 & 31.3/60.4 \\
AFC~\cite{AFC} (PR'24)& 82.7/93.6 & 40.7/70.5 \\
PPLR~\cite{pplr} (CVPR'22)& 84.4/94.3 & 42.2/73.3 \\
CCIOL~\cite{CCIOL} (PR'24)& 85.2/94.1 & 45.1/72.9 \\
ISE~\cite{ise} (CVPR'22)& 85.3/94.3 & 37.0/67.6 \\
HSP-MFL~\cite{HSP-MFL} (VCIR'23)& 85.5/95.3 & 45.2/75.1 \\
RTMem~\cite{RTMem} (TIP'23)& 86.5/94.3 & 38.5/63.3 \\
DCMIP~\cite{dcmip} (ICCV'23)& 86.7/94.7 & 40.9/69.3 \\
PASS~\cite{pass} (ECCV'22)& 88.5/94.9 & 41.0/67.0 \\
ACFL-VIT~\cite{ACFL} (PR'24)& \underline{89.1/95.1} & \underline{45.7/70.1} \\
TCMM (Ours) & \textbf{90.5}/\textbf{96.0} & \textbf{52.0}/\textbf{78.4} \\ \hline
\end{tabular}}
\label{tab:comparison}
\end{table}
%%%%%%%%%%%%%%%%%%%%%%%%%%%%%%%     Visualization    %%%%%%%%%%%%%%%%%%%%%%%%%%%%%%%%
\subsection{Visualization.}
\label{sec:Vis}
\begin{figure}[H]
\centering
\subfigtopskip=10pt  % 第一個子圖跟 caption 文字的距離

\subfigure[T-SNE visualization of features extracted by the TCMM model.]{
\label{fig:TCMM_tsne}
\includegraphics[width=\linewidth]{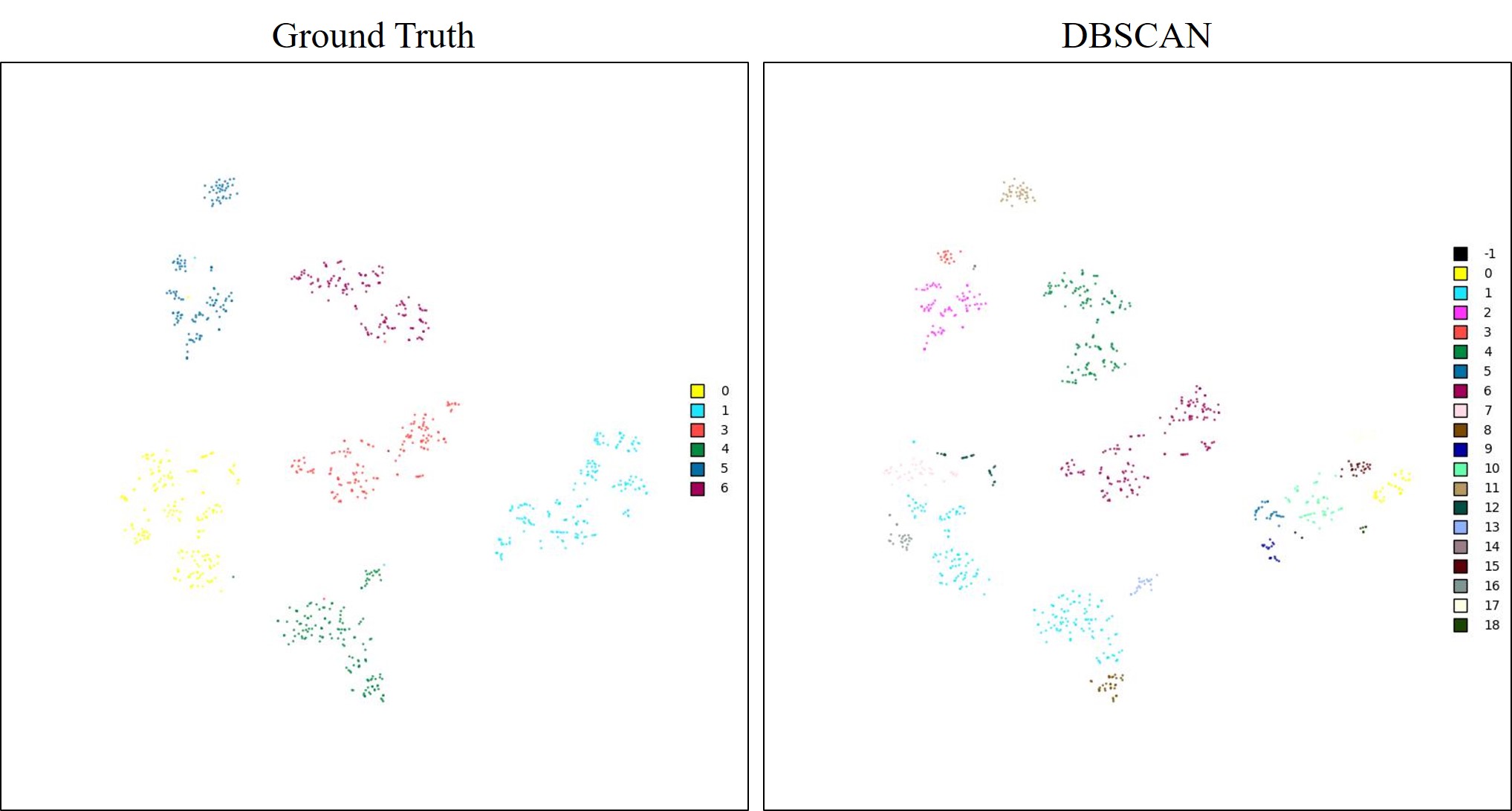}
}
\subfigure[T-SNE visualization of features extracted by the baseline model.]{
\label{fig:baseline_tsne}
\includegraphics[width=\linewidth]{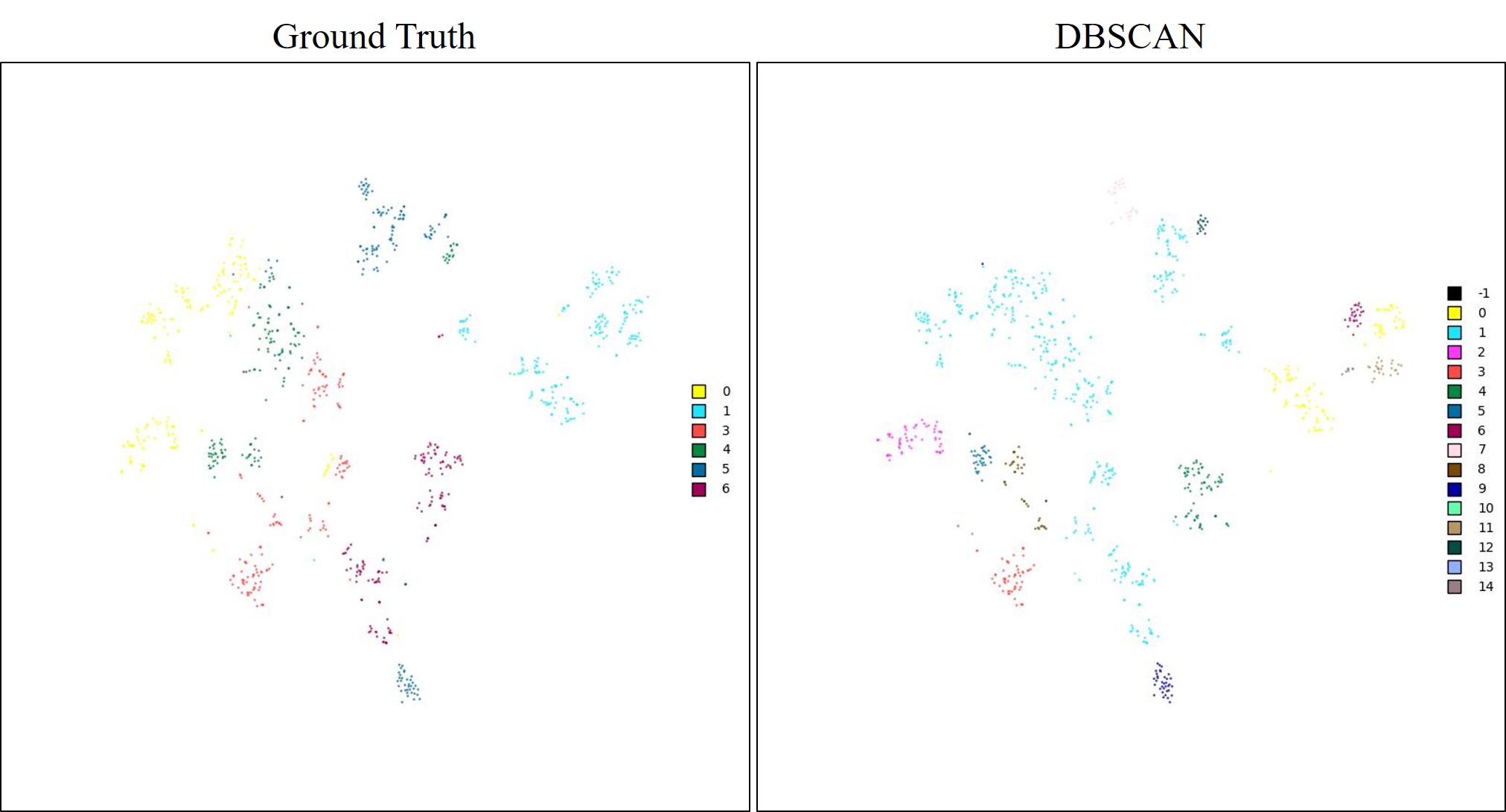}
}
\caption[t-sne visualization.]{The t-SNE visualization results on MSMT17 dataset. The same color represents samples classified under the same ID. The left side of Figure \ref{fig:TCMM_tsne} visualizes the features extracted by the TCMM model along with the ground truth, while the right side visualizes the clustering results using DBSCAN. Similarly, the left side of Figure \ref{fig:baseline_tsne} visualizes features extracted by the baseline model with the ground truth, and the right side shows the visualization of the clustering results using DBSCAN.}
\label{fig:tsne}
\end{figure}

To more intuitively demonstrate the effectiveness of TCMM, we conduct multiple visualization results.
In Figure \ref{fig:tsne}, we use t-SNE to visualize the learned features of the six identities with the most samples in the MSMT17 dataset. Points of the same color indicate that these samples belong to the same identity. Figure \ref{fig:TCMM_tsne} shows that the TCMM model enables features of the same identity to almost always cluster together, while images of different identities tend to be separated. In contrast, Figure \ref{fig:baseline_tsne} demonstrates that the baseline model more easily allows features of different identities to cluster together, making it harder to separate images of different identities.

\begin{figure}[t]
\centering
\includegraphics[width=\linewidth]{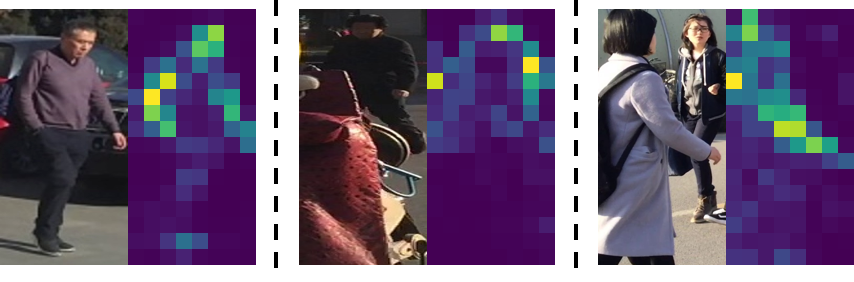}
\caption[attention vis.]{The attention maps visualization results of our TCMM. From left to right, the scenarios represent a general scene with background noise, a case with occlusions, and a situation with unrelated objects.}
\label{fig:attention}
\end{figure}

We also visualize attention maps to demonstrate our TCMM model performance. Figure \ref{fig:attention} presents the attention map results under three different scenarios. The result on the left shows that TCMM ignores background noise and focuses on the target for a given input image. The middle result shows that TCMM can ignore occlusions and focus on the target subject even under occluded conditions. The result on the right shows that even when an image contains two people, TCMM ignores unrelated objects and focuses on the correct target. From the Figure \ref{fig:attention}, it can be seen that TCMM can focus on the correct target in challenging scenarios, indicating that TCMM is capable of learning robust features.

\begin{figure}[h]
\centering
\includegraphics[width=\linewidth]{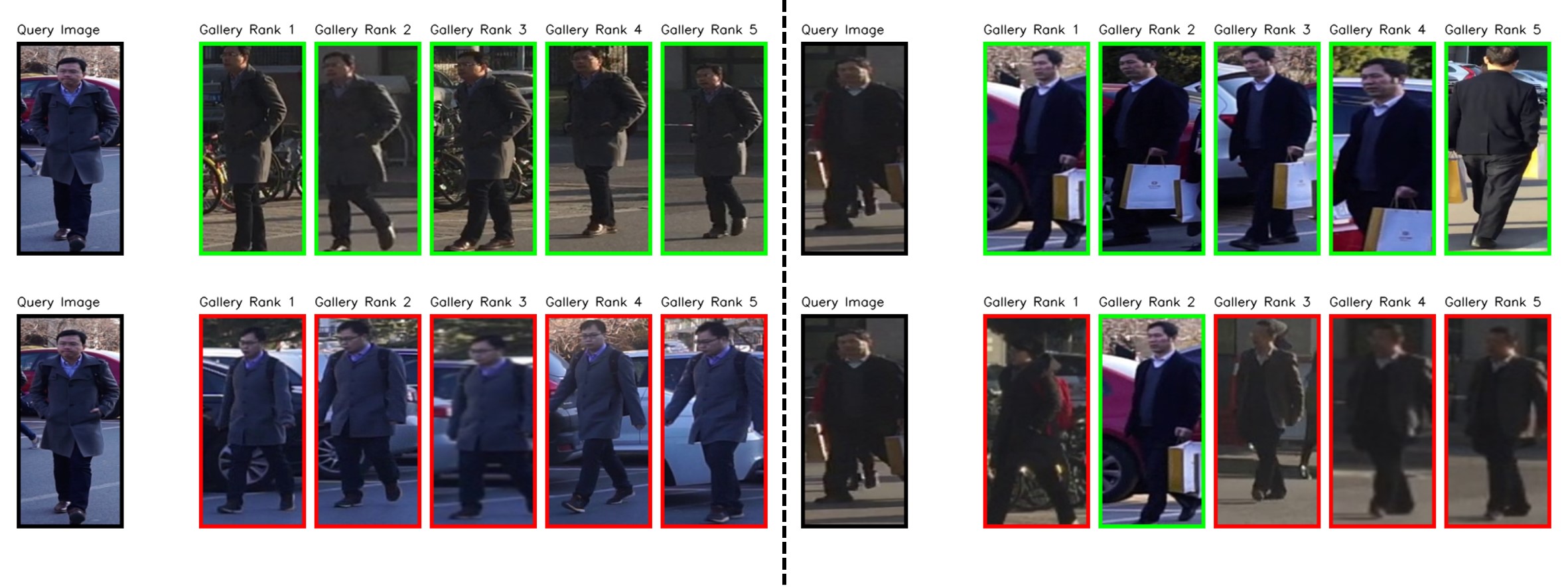}
\caption[ranking vis.]{The ranking list visualization results. The first row is the ranking list from the TCMM, and the second row is from the baseline. The green boxes indicate correct matching between the query image and images in the gallery set, while the red boxes denote incorrect matching.}
\label{fig:ranking}
\end{figure}

Finally, we compare the ranking results of our TCMM and the baseline on the MSMT17 dataset. As shown in Figure \ref{fig:ranking}, the green boxes indicate correct matching between the query image and images in the gallery set, while the red boxes denote incorrect matching. Figure \ref{fig:ranking} demonstrates that even when individuals wear similar clothing, TCMM can still learn discriminative features from diverse unlabeled training images. These examples validate the superior performance of TCMM over the baseline.

% ===========================  Conclusion  ===========================%
\section{Conclusion}

This paper introduces the ViT Token Constraint and Multi-scale Memory bank (TCMM) system to tackle existing unsupervised person re-id issues. The ViT token constraint imposes restrictions on ViT output tokens to alleviate the impact of patch noises on the ViT.
The multi-scale memory bank comprises the instance and the prototype level memory module. The instance memory encourages the model to learn from challenging samples, enabling it to cover various data variations and improve generalization capabilities. The prototype memory utilizes prototypes as positive and negative samples for contrastive learning to mitigate feature inconsistencies.
According to experimental results, TCMM achieves state-of-the-art performance on both Market1501 and the more complex MSMT17 datasets. This demonstrates that TCMM allows the model to learn more discriminative re-id features and enhances the robustness.
In the future, we hope to delve deeper into the self-attention mechanism to mitigate the damage caused by patch and pseudo label noises to the ViT.

\section*{CRediT authorship contribution statement}
\textbf{Zheng-An Zhu:} Conceptualization, Methodology, Software, Validation, Formal analysis, Investigation, Data Curation, Writing - Original Draft, Visualization. 
\textbf{Chen-Kuo Chiang:} Conceptualization, Resources, Writing - Review \& Editing, Supervision, Project administration, Funding acquisition.
\section*{Data availability}
Data will be made available on request.

\section*{Acknowledgements}
This work was funded by the National Science and Technology Council, Taiwan under Grant NSTC 112-2634-F-194 -001, NSTC 112-2634-F-006 -002 and NSTC 112-2927-I-194-001-.

% 35-55 篇
\bibliographystyle{elsarticle-num} 
\bibliography{reference}

% \begin{wrapfigure}{l}{25mm} 
%     \includegraphics[width=1in,height=1.25in,clip,keepaspectratio]{figures/Chen.Kuo.Chiang.jpg}
%   \end{wrapfigure}\par
%   \textbf{Chen-Kuo Chiang} Chen-Kuo Chiang is an assistant professor at the Department of Computer Science and Information Engineering, National Chung Cheng University, Taiwan. He received his Ph.D degree in computer science department at National Tsing Hua University, Hsinchu, Taiwan. His research interests include computer vision, machine learning, pattern recognition and image processing. \par
  
% \begin{wrapfigure}{l}{25mm} 
%     \includegraphics[width=1in,height=1.25in,clip,keepaspectratio]{figures/ZHUZHENGAN.jpg}
%   \end{wrapfigure}\par
%   \textbf{Zheng-An Zhu} Zheng-An Zhu is a Ph.D student at the Department of Computer Science and Information Engineering, National Chung Cheng University, Taiwan. His research interests include computer vision, machine learning and pattern recognition.\par

\end{document}